\documentclass[letterpaper, 10 pt, conference]{ieeeconf}
\IEEEoverridecommandlockouts
\usepackage{times}

\usepackage{graphicx}
\graphicspath{{figures/}}

\usepackage{amsmath}

\usepackage{amssymb}
\usepackage{tabularx}
\usepackage[utf8]{inputenc}
\usepackage[T1]{fontenc}
\usepackage{textcomp}
\usepackage{gensymb}
\usepackage{multirow}

\usepackage{float}
\usepackage{algorithm}
\usepackage[noend]{algpseudocode}
\newfloat{algorithm}{t}{lop}

\usepackage{cite}

\usepackage[printonlyused]{acronym}

\acrodef{RL}{Reinforcement Learning}
\acrodef{DRL}{Deep Reinforcement Learning}
\acrodef{DR}{Domain Randomization}
\acrodef{ML}{Markov Localization}
\acrodef{AMCL}{Adaptive Monte Carlo Localization}
\acrodef{DAL}{Deep Active Localization}
\acrodef{ANL}{Active Neural Localization}
\acrodef{AML}{Active Markov Localization}

\newcommand{\Mm}{\mathcal{M}}
\usepackage{ntheorem}
\newtheorem*{problem}{Problem}

\DeclareMathOperator*{\argmax}{\arg\!\max}

\newcommand{\lidar}{LiDAR}

\usepackage{multicol}
\usepackage[colorlinks=true,bookmarks=true]{hyperref}
\usepackage{subfigure}
\usepackage[usenames,dvipsnames]{xcolor}
\usepackage{capt-of}
\usepackage{caption}

{

}

\begin{document}

\title{\textbf{Deep Active Localization}}

\author{Sai Krishna$^{1*}$, Keehong Seo$^{2*}$, Dhaivat Bhatt$^{1}$, Vincent Mai$^{1}$, Krishna Murthy$^{1}$, Liam Paull$^{1}$
\thanks{$^1$ Mila and DIRO, Universite de Montreal, Canada}
\thanks{$^2$ SAIT, Samsung Electronics Co. Ltd. Suwon, South Korea}
\thanks{$^*$ The first two authors contributed equally to this work. }
%
}

\thispagestyle{plain}
\pagestyle{plain}

\maketitle

\begin{abstract}
\label{abstract_lb}
  Active localization is the problem of generating robot actions that allow it to maximally disambiguate its pose within a reference map. Traditional approaches to this use an information-theoretic criterion for action selection and hand-crafted perceptual models. In this work we propose an end-to-end differentiable method for learning to take informative actions that is trainable entirely in simulation and then transferable to real robot hardware with zero refinement. The system is composed of two modules: a convolutional neural network for perception, and a deep reinforcement learned planning module. We introduce a multi-scale approach to the learned perceptual model since the accuracy needed to perform action selection with reinforcement learning is much less than the accuracy needed for robot control. We demonstrate that the resulting system outperforms using the traditional approach for either perception or planning. We also demonstrate our approaches robustness to different map configurations and other nuisance parameters through the use of domain randomization in training. The code (available \href{https://github.com/montrealrobotics/dal}{here}) is compatible with the OpenAI gym framework, as well as the Gazebo simulator.

\end{abstract}

\maketitle

\section{Introduction}
\label{intro_lb}
\emph{Localization} against a global map, or \emph{global} localization, is a prerequisite for any robotics task where a robot must know where it is (e.g. any task involving navigation). In such settings, it is beneficial for the agent to first select actions in order to disambiguate its location within the environment (map). This is referred to as ``\emph{active} localization''.


Traditional methods for localization, such as \ac{ML} \cite{markov_localization} and \ac{AMCL} \cite{amcl} are ``\emph{passive}'' (agnostic to how actions are selected). They provide a recursive framework for updating an approximation of the state belief posterior as new measurements arrive. In the case of \ac{ML}, this is usually done by some type of discretization of the state space (i.e. a fixed grid) and in \ac{AMCL} is achieved by maintaining a set of \emph{particles} (state hypotheses). While these methods have seen widespread success in practice, they are still fundamentally limited since the map representation is hand-engineered and specifically tailored for the given on-board robot sensor. A common example is the pairing of the occupancy grid map \cite{Elfes} with the laser scanner since the measurement likelihood can be computed efficiently and in closed form with scan matching \cite{scan_matching}. However, this choice of representation can be sub-optimal and inflexible. Furthermore, sensor parameters such as error covariances tend to be hand-tuned for performance, which is time consuming and error-inducing.


Active variants of \ac{ML} and \ac{AMCL} are classical ways to solve the active localization problem \cite{active_seminal,thrun1998finding,arbel1999viewpoint, active_ml,kummerle08}. These methods typically leverage an information-theoretic measure to evaluate the benefit of visiting unseen poses.  Again, these hand-tuned heuristics tends to lead to over-fitting of a particular algorithm to a specific robot-sensor-environment setup. 

\begin{figure}[!t]
    \centering
    \includegraphics[width=0.8\columnwidth]{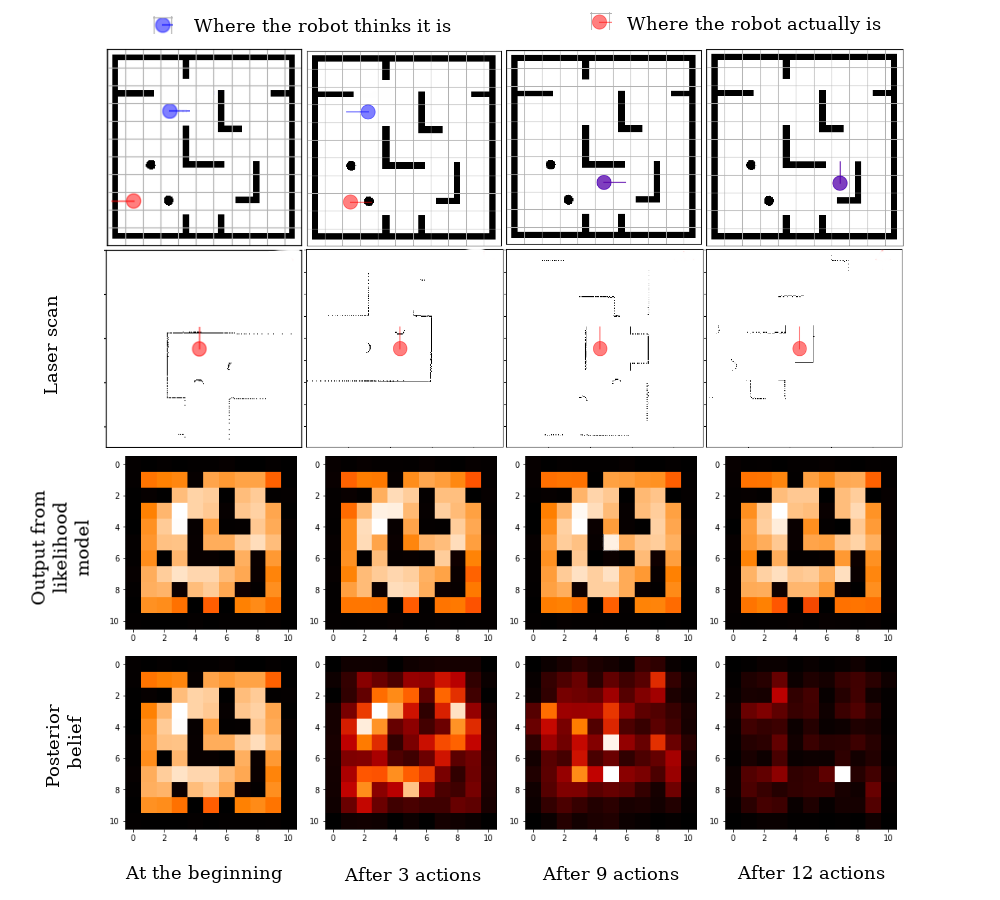}
    \caption{DAL demonstration in simulation. At the beginning, the robot has no real clue of where it is relative to a known map, but as it executes actions and moves around, it converges onto a true location (see posterior belief improving in the fourth row). Furthermore, it transfers \emph{zero-shot} onto a real robot, outperforming several baseline active localization methods. Refer to Sec. \ref{sec:results} for more details. \vspace{-0.1cm}}
    \label{fig:teaser}
\end{figure}

Passive deep learning localization algorithms \cite{posenet,Valada18icra,DPF1,DPF2} are not directly extendable to the active localization domain in the same fashion because they tend not to output calibrated measures of uncertainty.

However, other learning-based methods have been built that are dedicated specifically to this task, such as \ac{ANL} \cite{ANL}. In the \ac{ANL} framework, a \ac{DRL} policy model is trained on the outputs of a \emph{fixed} measurement likelihood model (no learnable parameters), but still within a recursive Bayesian framework. However, several of the assumptions made in this work simply do not apply in real world robot deployments (for a more detailed analysis please see Sec. \ref{sec:related}).

\ac{DRL} has seen an incredible recent success on some robotics tasks such as manipulation \cite{Levine_Deep_policy_15} and navigation \cite{drl_navigation}, albeit predominantly in simulation. In order for a \ac{DRL} agent to be trained in simulation and deployed on a real robot either 1) the reality gap is small or is overcome with training techniques such as \ac{DR} such that no refinement of the policy is needed on the real robot (zero-shot transfer), or 2) the agent policy is fine-tuned on the real robot after primarily training in simulation. The latter option reduces the burden on the simulation fidelity and training regime, but fine-tuning on the robot can be difficult or impossible if the reward is determined by leveraging ground truth parameters (like localization) that are only available in the simulator.   
In this work, we argue that the framework of \ac{DRL} holds potential for the task of active localization, and can be used to train policies that transfer to a robot in real-world indoor environments. Borrowing concepts from Bayes filtering \cite{probabilistic_robotics}, we define a reward as a function of the posterior belief. Specifically, our reward is defined as the probability of posterior belief at ground truth pose. 




\subsection{Contributions}
\label{contrib_lb}
In summary we claim the following contributions:

\begin{itemize}
\item We propose a multi-layer learned  likelihood model which can be trained in simulation from automatically labeled data and then refined in an end-to-end manner,
\item We show that this method  works for \emph{zero-shot} transfer onto a real robot and outperforms classical methods.
\item We have integrated classical (Bayesian filtering, scan matching for ground truth likelihood) and learning-based (supervised and reinforcement learning) approaches appropriately to yield an active localization system that runs on a real robot.
\item We developed automated processes for map generation, domain randomization and likelihood and policy model training.
\item We provide an openAI-gym environment for the task of active localization to facilitate further research in this area. We have integrated it with SOTA deep RL algorithms to provide baselines
\item We outline a set of domain randomization techniques and show that our learned likelihood model is more robust than classical hand-tuned techniques.
\end{itemize}

\section{Overview of Approach}


\begin{figure*}
    \centering
    \includegraphics[width=15cm, height=6cm]{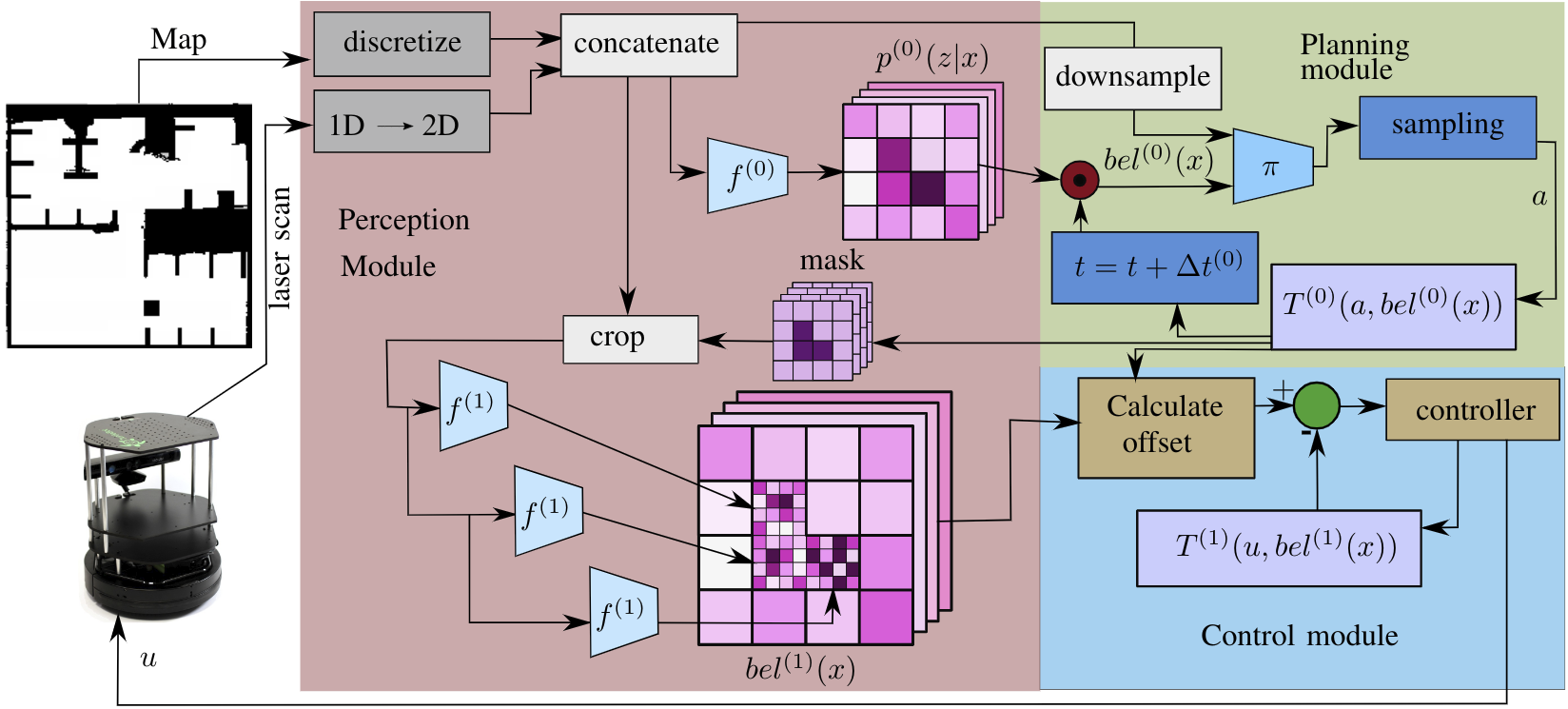}
    \vspace{0.1cm}
    \caption{\textbf{Deep Active Localization}: Our proposed method takes a map and sensor data (laser scan) and generates control actions in an end-to-end differentiable framework that includes learned perceptual modules (neural networks) at different scales and a learned policy network that is trained with reinforcement learning (RL). We maintain a coarse low dimensional pose estimate for RL, but then refine it to generate a higher precision pose estimate that is used for robot control.
    \vspace{-0.1cm}}
    \label{fig:system}
\end{figure*}

In this section we will define our problem precisely and outline the structure of our proposed solution.

\begin{algorithm}[t!]
\caption{Deep Active Localization}
\label{alg:dal}
\begin{algorithmic}[1]
\Procedure{DAL}{$\Mm, f^{(0)}, f^{(1)}, \pi$}
\State \textbf{Input}:
\State $\Mm$: Map of the environment
\State $f^{(0)}$: Trained likelihood model for level-0
\State $f^{(1)}$: Trained likelihood model for level-1
\State $\pi$: Trained policy model
\While{True}
\State Read \lidar{} scan $z_t$ 
\State Obtain high and low res scan images $S_h$ and $S_l$ 
\State $p^{(0)}(z_t|x) \gets f^{(0)}(M_h, S_h)$ \Comment{Low-res likelihood}
\State $bel^{(0)} \gets p^{(0)}(z_t|x) \odot \bar{bel^{(0)}}$ \Comment{Update belief}
\State $p(a|s) \gets \pi(bel^{(0)}, M_l, S_l)$ \Comment{Get action prob}
\State $a \gets sample(p(a|s))$ \Comment{Sample an action}
\State $c_t \gets arg max(bel^{(0)})$ \Comment{Get current pose}
\State $g_t \gets$ get-next-pose($c_t, a$) 
\State $\bar{bel^{(0)}} \gets$ transition-belief($bel^{(0)}, a$)

    
    \State find top $s$ cells in $p^{(0)}(z_t|x)$
    \For{\texttt{$i \in \left[0,s\right]$}}
        \State $M_c$, $S_c \gets$ crop($M_h$, $S_h$, $(n_i, m_i)$)
        \State $p^{(1)}(z_t|x)[block_i]\gets f^{(1)}(M_c, S_c)$
    \EndFor
    \State $bel^{(1)} \gets p^{(1)}(z_t|x) \odot \bar{bel^{(1)}}$ \Comment{High-res belief}
    \State $c_t^{(1)} \gets arg max(bel^{(1)})$  \Comment{Pose: max belief}
    \State $u \gets control(c_t^{(1)}, g_t)$
\EndWhile\label{euclidendwhile}
\EndProcedure
\end{algorithmic}
\end{algorithm}

\subsection{Problem Setup}
\label{sec:prob_def}
A mobile robot is equipped with a sensor that provides exteroceptive measurements $z_t$ used for localization and proprioceptive measurements used for odometry $o_t$ (this could be the same physical sensor). The robot moves through the map by sending control inputs $u_t$ to its actuators. The agent is provided with a map $\Mm$.

\begin{problem} (\textbf{Active Localization})
\label{pr:al}
  Assuming that the agent is placed at some point in the map, find the sequence of control inputs, $u_{1:T}$ that allow it to maximally disambiguate its pose within the map.
\end{problem} 
\vspace{0.2cm}

Active localization solutions usually take the form:

\begin{equation}
  u^*_{1:T} = \argmax_{u_{1:T}}f(bel(x_T),x^*)
  \label{eq:al}
\end{equation}
where the function $f$ in some way quantifies the weight in the state belief posterior at the end of the horizon $T$, $bel(x_T)$ at the ground truth pose $x^*$.

Any algorithm used to solve the optimization in \eqref{eq:al} must jointly consider both the \textit{perception} task (i.e., how the belief posterior is being updated) as well as the planning and control problem that is used to generate the control actions conditioned on the belief.

\subsection{System Overview}
\label{subsec:system-overview}

Our approach to solving the active localization problem is summarized in Fig.~\ref{fig:system} and Algorithm \ref{alg:dal} \footnote{Conventions: superscripts $x^{(i)}$ denote levels in the hierarchy, subscripts denote time indices, and preceding superscripts ${}^{\{r\}}x$ denote reference frames. In the absence of a frame it is assumed that the variable is in the map fixed (global) frame.}. For simplicity, we have shown two hierarchical levels in Fig.~\ref{fig:system} since it is appropriate for our setup, but further levels of hierarchical could be added following the same approach(i.e., by providing: a measurement likelihood model, a planner, and a transition function)
At each level in the hierarchy, $i$, we define a grid representation of size $N^{(i)}\times M^{(i)} \times \Theta^{(i)}$ which represents the state space $\mathcal{X}$. The belief posterior is represented as a matrix where $bel(x_t=[n,m,\theta])$ is the probability mass at location $[m,n,\theta]$. 

In Fig.~\ref{fig:system}, we assume that the neural network models are already trained, for a description of the training procedure for the measurement likelihood models see Sec.~\ref{sec:perception} and for the \ac{RL} policy see Sec.~\ref{sec:policy}.

The robot is provided with a map, $\Mm$ as input. Each time a sensor input, $z_t$ is received, it is converted to a 2D image and, combined with the $\Mm$  to form the input to the network at level 0, $f^{(0)}(\Mm,z_t)$, which produces a coarse measurement likelihood, $p^{(0)}(z_t|x)$. The measurement likelihood is combined with the prior belief (via element-wise product) to produce the belief posterior. The belief posterior is fed as input to the \ac{RL} model. The other inputs to the \ac{RL} model are low dimensional map and low dimensional scan. It is important in order for this procedure to train efficiently that the input is relatively low dimensional. The \ac{RL} policy is sampled to generate an action, $a_t$ from the set $\{$ \texttt{left}, \texttt{right}, \texttt{straight} $\}$. This action produces a goal pose that is either the centroid of an adjacent cell at the same orientation, or a pure rotation of $\pm 2\pi/|\Theta^{(0)}|$, where $\Theta^
{(0)}$ is the number of uniformly spaced discrete angles being considered at level 0. The action and belief posterior are fed through a noisy transition function to generate the belief prior at the next timestep:

{\small
\begin{equation}
  \begin{split}
    & \bar{bel}^{(0)}(x_{t+1}=[n,m,\theta]) = T^{(0)}(a_k, bel(x_t))  \\
    & = \begin{cases}
      bel^{(0)}(x_{t}=[n,m,(\theta - 2\pi/|\Theta^{(0)}|)] & a_k=left \\
      bel^{(0)}(x_{t}=[n,m,(\theta + 2\pi/|\Theta^{(0)}|)] & a_k=right \\
      bel^{(0)}(x_{t}=[n+l_n\cos(\theta),m+l_m\sin(\theta),\theta]) & a_k=str
    \end{cases}
  \end{split}
\end{equation}}
where $l_n$ and $l_m$ are the distance between adjacent cell centroids in the $N^{(0)}$ and $M^{(0)}$ directions respectively. For increased robustness, noise is injected into this transition model to represent the fact that the cell transition is not deterministic.

Since each time the robot executes an action it will not arrive exactly at the centroid of the adjacent cell due to dead reckoning error, we must account for the error accrued and compensate for it. We refine our coarse measurement likelihood to predict where \textit{within} the cell our agent actually is to compute a correction. This offset within the cell is used to calculate an offset for the next action so that the error with respect to the grid centroids is \textit{bounded over time}. To achieve this, we chose the ``likely'' cells from the coarse belief $bel^{(0)}(x)$ and refine the measurement likelihood to determine a more precise estimate of the location of the agent within the cell. This is used to calculate an offset to add to the relative pose transformations between adjacent cells to calculate an actual reference location in the robot frame, $r$: ${}^rx_{ref}$. We use a standard tracking controller to generate control inputs, $u_t$, which are sent to the robot and a standard kino-dynamic transition model $T^{(1)}(u,bel^{(1)}(x))$ to dead-reckon towards the reference location.

\section{Perception}

The objectives of the perception system are:
\begin{enumerate}
\item To provide a measurement likelihood to be used by \ac{RL} agent, which is trainable in an end-to-end manner,
\item  To provide a refined estimate of pose to the inner loop controller so that the error induced by dead reckoning may be bounded,
\item To be fully trainable in simulation,
  \item Not to require any information other than the (possibly noisy) map at test time.
\end{enumerate}

\subsection{Learning Measurement Models}
\label{sec:perception}
In typical robotics pipelines, the measurement likelihood is constructed using custom metrics, based on models of the sensors in question. Inevitably, some elements of the model are imprecise. For example, covariances in visual odometry models are typically tuned manually since closed-form solutions are difficult to obtain. Additionally, the Gaussian assumption (e.g., in an EKF) is clearly inappropriate for the task of global localization, as evidenced by the prevalence of Monte Carlo based solutions \cite{amcl}.

In our setting, given a map of the environment $\Mm$ and scan input $z_t$, we want to \emph{learn} the likelihood of the robot's pose at all candidate points on a multi-resolution grid.

Following the notation defined in Sec \ref{subsec:system-overview}, the output of the likelihood model at time step $t$ is 
given by:

\begin{equation}
p^{(0)}(z_t | x) = f^{(0)}_{\phi}(\Mm,z_t)
\end{equation}
where $\phi$ are the parameters of the neural network model.

\subsubsection{Data Preprocessing}

It is difficult for a neural network to learn from different dimensional inputs. Though we can use different embeddings and concatenate at deeper layers, our experiments revealed that this is not very efficient. So, as a pre-processing step, we convert the scan input $z_t$ into a scan image $S_h$ ($h$ denotes high resolution) of the same size as the grid map. We concatenate both the 2D scan image and the map of the environment and use it as input to our neural network. During the training phase, we obtain the ground truth likelihood $\tilde{p}^{(0)}(z_t|x)$ by taking the cosine similarity or correlation of the current scan with the scans at all other possible positions.

\subsubsection{Ground Truth Data Generation}

We generate and store triplets of (scan-image $S_h$, map of the environment $M_h$, ground truth likelihood $\tilde{p}^{(0)}(z_t|x)$) while randomly moving the robot in the simulator and randomly resetting the initial pose after every $T$ time steps and randomly resetting the environment after every $e$ such episodes. We train it on $m$ such environments (maps). 
Training on these triplets using Resnet-152 \cite{resnet} or Densenet-121 \cite{densenet} gave very good results in simulation but did not transfer when these models are transferred to real robot in a real environment. This demonstrates the need for domain randomization to achieve zero-shot transfer. See \ref{subsec:domain-randomization} for further details. 

\subsection{Bayes' Filter}

We can now update the belief by taking element wise product of likelihood and the previous belief over all the $N \times M \times O$ grid cells
$$bel^{(0)} = \bar{bel}^{(0)} \odot p^{(0)}(z_k|x)$$

\subsection{Hierarchical Likelihood Model}
\label{subsec:hierarchical-lieklihood}

We estimate the robot pose on a coarse grid, which is sufficient for RL, but we require a refined estimate for two reasons:

\begin{enumerate}
    \item It's possible that the coarse estimate is simply not precise enough, particularly in very large maps
    \item Without a more precise estimate of the robot's pose within a coarse grid cell, the robot will gradually drift from the grid centroids as a result of dead reckoning.
\end{enumerate}

Traditional approaches try to solve this problem using scan matching technique, but it has the drawback of being computationally very expensive as the current scan information has to be compared with scan information at multiple poses. 

Recent approaches like ANL\cite{ANL} try to overcome this problem by using neural networks to predict the likelihood in a 3D grid of dimensions $N^{(1)} \times M^{(1)} \times \Theta^{(1)}$, where the whole environment is divided into $N^{(1)}$ rows, $M^{(1)}$ columns and $\Theta^{(1)}$ orientations. Each of these grid cells represent the likelihood of robot in that pose. But, this doesn’t alleviate the problem of decoupling the localization precision from the size of the map. With input being $N \times M \times 2$, and the output being $N \times M \times \Theta$, it is usually difficult and inefficient for a CNN to learn this mapping. 

In HLE (Hierarchical Likelihood Estimation), likelihood is first estimated at coarse resolution ($N^{(0)} \times M^{(0)} \times O $) and then each of these grid cells is expanded to further finer grid cells ($k \times k$) to get a full resolution map of size ($N^{(1)} \times M^{(1)} \times \Theta$).

For the case of 2-level hierarchy, we have 2 neural networks $f^{(0)}(M_h,S_h)$ and $f^{(1)}(M_c, S_c)$ at levels 0 and 1 respectively.
The input for $f^{(0)}$ is high the resolution map $M_h$ (which is a processed version of map $\Mm$ of the environment) ($N^0\times M^0 $), and scan image $S_h$ ($N^0 \times M^0 $) which is obtained from the current \lidar{} scan $z_t$  at the current pose. 
The likelihood output $p^{(0)}(z_t|x) = f^{(0)}(M_h,S_h)$ is of shape $N^{(0)} \times M^{(0)} \times \Theta^{(0)}$. The parameters of $f^{(0)}$ are optimized by minimizing the mean squared error (MSE) between $p^{(0)}(z_t|x)$ and ground truth likelihood $\tilde{p}^{(0)}(z_t|x)$ (which is at the same resolution of $N^{(0)}\times M^{(0)} \times \Theta^{(0)}$). 

Grid cells corresponding to a maximum of $c$ values from the coarse likelihood $p^{(0)}(z_t|x)$ are selected. For each of these grid cells, we crop a square patch from the high resolution map and high resolution scan (optional) around the grid cell. Concatenation of the cropped map and cropped scan is used as input for $f^{(1)}$. The network outputs a $k \times k$ block where each cell in the block represents likelihood at finer resolution. each of the cell in this block is multiplied with corresponding likelihood of the grid cell at previous level $p^{(0)}(z_t|x)$.
For the cells which are not in the top $c$ values, the likelihood value at level-0  $p^{(0)}(z_t|x)$ is normalized and directly copied to each cell in the corresponding block in $p^{(0)}(z_t|x)$. The parameters of $f^{(1)}$ are optimized by minimizing the mean squared error (MSE) between $p^{(1)}(z_t|x)$ and ground truth likelihood $\tilde{p}^{(1)}(z_t|x)$ (which is at the same resolution of $N^{(1)}\times M^{(1)} \times \Theta^{(1)}$).

\subsection{Domain Randomization}
\label{subsec:domain-randomization}

\ac{DR} has become a popular method to promote generalization, particularly when training an agent in a (necessarily imperfect) simulator, and then deploying it on real hardware \cite{Tobin}. 
While the \lidar{} data modality  generally transfers better that vision (camera images) from simulation to the real world scenario 
it is not without challenges since no sensor model is perfect. Just training the likelihood models without any variability in the simulator results in over-fitting and poor transfer. Since we do not have an exact recreation of the real world environment within our simulator, our learned agent will have to generalize to a new environment while simultaneously bridging the reality gap. Hence, it is important to account for various real world irregularities while training on the simulator. In our data collection pipeline, we randomized the following parameters:

\begin{itemize}
\item Thickness and length of obstacles to account for different types of obstacles in real environment.
\item Error in robot pose to account for the possibility that it is not exactly at the centroid of a given cell.
\item Temperature of softmax.
It is important to normalize the ground truth likelihood to keep it bounded for a machine learning system to be able to learn. We use softmax with temperature $\beta$ defined as: 
\begin{equation}
    \sigma(p^{(0)}(z_t|x))_{n,m,\theta} = \frac{e^{\beta p^{(0)}(z_t|x)_{n,m,\theta}}}{\sum_{n,m,\theta} e^{\beta p^{(0)}(z_t|x)_{n,m,\theta}} }
\end{equation}
We vary the temperature parameter $\beta$ randomly within limits to make sure that it is not biased towards overly-uniform or overly-sharp likelihood outputs

\item Noise in the \lidar{} scan: We train our network by adding Gaussian noise to every \lidar{} scan point. Additionally, for every scan, we randomly set some of the incoming data points within the scan to have the value of $+\infty$. 

\item There are errors in the actual map of the environment created using gmapping \cite{gmapping}. This map is preprocessed and used as an input to the likelihood model and RL model for experiments on the real robot. Hence, it is important to ensure that some noise (more erosion and dilation of the map) is added to the map during training phase (on simulation) as well. However, note that the \lidar{} readings will still come from the unperturbed map.   

\end{itemize}

\section{Planning and Control}

The multi-scale localization estimates are used for planning and control.

\subsection{Reinforcement Learning}
\label{sec:policy}

\begin{figure}
    \centering
    \includegraphics[width=\columnwidth]{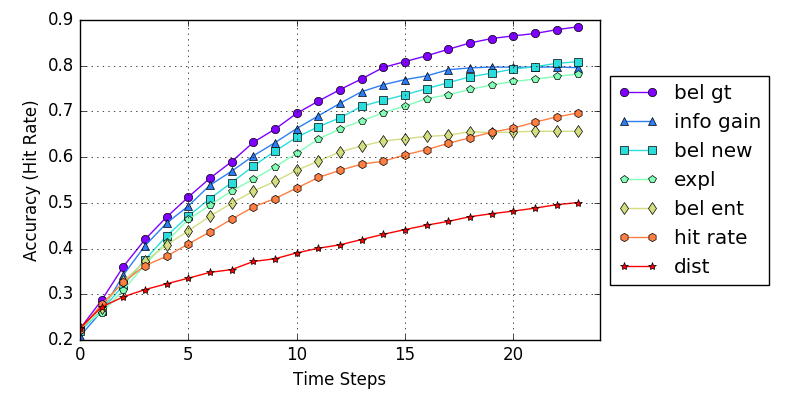}
    \vspace{-0.6cm}
    \caption{We trained 7 RL policy models with different rewards for each: probability mass of belief at true pose (bel gt), decrease in the entropy of the belief (info gain), reward of +1 for a new pose in belief (bel new), reward of +1 for a new true pose (expl), negative entropy of belief (bel ent), reward of +1 if Manhattan distance error is equal to 0 (hit rate),  Manhattan distance error (dist). 
    \vspace{-0.1cm}}
    \label{fig:reward-compare}
\end{figure}

Following the traditional conventions in a Markov Decision Process \cite{Sutton97}, we denote the state $ s \in \mathbb{R}^d $ and the actions $ a \in \mathbb{R}^{d_a} $, the reward function $r(s,a)$ and a deterministic transition model $T = p(s' | s,a)$. In our case, the state (input to the RL model) is a concatenation of belief map $bel(x_t)$ ($N^{(0)} \times M^{(0)} \times \Theta^{(0)}$) and the input map of the environment $\Mm$. 
We formulate the MDP over high-level actions (\texttt{left}, \texttt{right}, and \texttt{straight}) 
The goal of any RL algorithm is to optimize it's policy $\pi(a|s)$ to maximize the discounted return defined as: $G_t = R_{t} + \gamma R_{t+1} + ... $ over initial distribution of states (which is assumed uniform here). We use advantage actor critic (A2C) \cite{Mnih2016} algorithm to accomplish this task. There are various choices for a reward function. We used belief at true pose as our reward function since it is dense, well-behaved, and benefits from the availability of true pose in the simulator. The empirical evidence for the choice is given in Fig-\ref{fig:reward-compare}.

\subsection{Closed-Loop Control}

After the high level actions (\texttt{left}, \texttt{right}, \texttt{straight}) are chosen by the RL model, it is important to ensure that the robot reaches it's goal position without deviating from the path. Even a minor deviation in each time step results in compounding errors. So, the low level actions (linear and angular velocities of the mobile robot) are given based on the current pose and the goal pose. The current pose is obtained at much finer resolution using our hierarchical likelihood model. See the Sec. \ref{subsec:hierarchical-lieklihood} for more details of hierarchical models. Optionally, the HLE model can be used after every fixed number of time steps to correct for the accumulated drift.

\subsection{Zero-shot Transfer}

The transferability of the perceptual model is enabled through the use of domain randomization (DR) as discussed in Sec. \ref{subsec:domain-randomization}.
Another common use of DR is to randomize over physical properties of the robot (i.e., dynamics). This is not necessary in our case since we are performing RL at the \emph{planning} level of abstraction and using a more traditional feedback controller to execute the plans. 

\section{Experiments}
\label{sec:results}

We demonstrate the efficacy of DAL by conducting several experiments in simulation and on real robots and varying environments. This section describes the experimental setup and presents the results obtained, which demonstrate that DAL outperforms state-of-the-art active localization approaches in terms of localization performance and robustness.


The real environments used for testing were simple and re-configurable. Using a SLAM software we then obtained maps as shown in Fig. \ref{fig:exp_maps}. 

\subsection{Dataset Generation}
Using the following algorithm, we generated a dataset to train the likelihood model to estimate the measurement likelihood $p(z|x)$. The algorithm generates random maps, sensor data at some random poses as well as the target distributions at those poses.
\begin{enumerate}
    \item We use Kruskal's algorithm to generate random maze-like environments at low resolution. The output is a square matrix with $1$s representing obstacles and $0$s representing free cells. The algorithm guarantees that all the open spaces are connected. 
  \item Based on the grid map at low resolution generated above, we place obstacles (of different sizes) at the cells in the high resolution. The algorithm then randomly dilates and erodes the surfaces of obstacles in the map so that its texture becomes more realistic. 

\item  We place the robot at all the grid centers of low resolution map at all $\Theta^{(0)}$ directions. From each of those locations, we simulate a laser scan. So, we obtain $M^{(0)}\times N^{(0)} \times \Theta^{(0)}$ laser scans. We call this as scan matrix and it's of dimension $M^{(0)}\times N^{(0)} \times \Theta^{(0)} \times 360$. 
Given a map, we precompute a range vector at each location with respect to the low-resolution grids. A range vector in our experiment has the length of 360, representing distances at each of 360 degrees.

\item We now spawn the robot at 10 random locations and compute the cosine similarity of the scan at the current pose with the scan at all possible poses to get likelihood map.  With each of those likelihoods, we form a triplet of {map, current scan, likelihood} and use these as training set

\end{enumerate}

The procedure was repeated for 10000 different maps. The noise added to this process is described in \ref{subsec:domain-randomization} and \ref{subsec:training-experiments}

\begin{figure}
    \centering
    \includegraphics[width=\columnwidth]{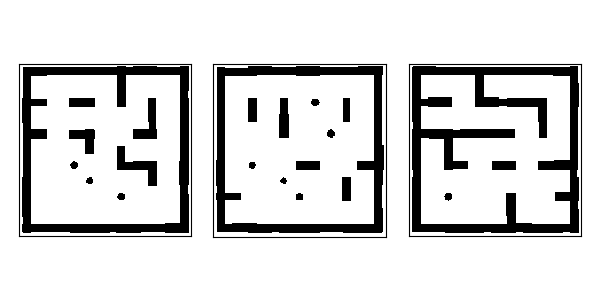}
    \caption{Random maps generated from Kruskal's maze algorithm with some of the walls pruned 
    }
    \label{fig:random_maps}
\end{figure}

\subsection{Training Likelihood and Policy Models}

\label{subsec:training-experiments}

We first describe how we trained our likelihood model (LM) which was used in our real robot experiments. We generated a dataset containing 10,000 maps with 10 scan inputs for each map by following the procedure explained above. The dimension of ground truth likelihood (GTL) was set to $33 \times 33 \times 8$, i.e, 8 headings with 33*33 $x$ and $y$ locations.  

Likelihood model was trained with a densenet201 model on this dataset. We applied the following randomization for training LM:
\begin{itemize}
    \item the temperature for softmax function was manually decreased over epochs from 1.0 to 0.1,
    \item input scan was randomly rotated in the range of $\pm2\degree$,
    \item 100 randomly selected pixels were flipped ($0 \leftrightarrow 1$),
    \item and drop-out was applied in densenet with the rate of 0.1.
\end{itemize} 


Likelihood model was trained first with 100,000 data instances over 16 epochs and adapted to the map of the real environment in a simulator for 10,000 inputs.

The policy model $\pi$ was trained on a simulator with randomly generated 2,000 maps, with 20 episodes (each of length 24) for each map. Reward (equal to probability mass of belief at true pose) was given at each time step.




\subsection{Experimental Setup}









To demonstrate the feasibility and robustness of our approach, we tested our trained likelihood and policy models on 2 mobile robots: JAY and Turtlebot and successfully localized in two different environments.

\subsubsection{Experiments on JAY}

JAY (Rainbow Robotics) is cylinder-shaped with the radius of 0.30m, the height of 0.50m and the weight of 46 kg before modification, whose CPU is IntelCo re i7-8809G Processor.
We mounted an extra \lidar{} A2M8 (Slamtec) on the top of the robot in the center to secure 360-degree view. The \lidar{} provides 0 to 360 degree angular range, and 0.15 to 8.0 m distance range, with 0.9 degree angular resolution, at the rate of 10 Hz.

The ROS navigation package including  AMCL was manually initialized and  used for ground truth localization. DAL was running on a separate server equipped with 4 Nvidia GTX-1080Ti GPUs and an Intel CPU E5-2630 v4 (2.1 GHz, 10 cores) to send velocity commands to JAY over a 2.4 GHz WIFI channel. 

\begin{figure}[!hbt]
    \centering
    \subfigure[JAY with top-mounted 360-degree view \lidar{}
    \label{fig:jay_intro}]{
    \includegraphics[width=0.4\columnwidth]{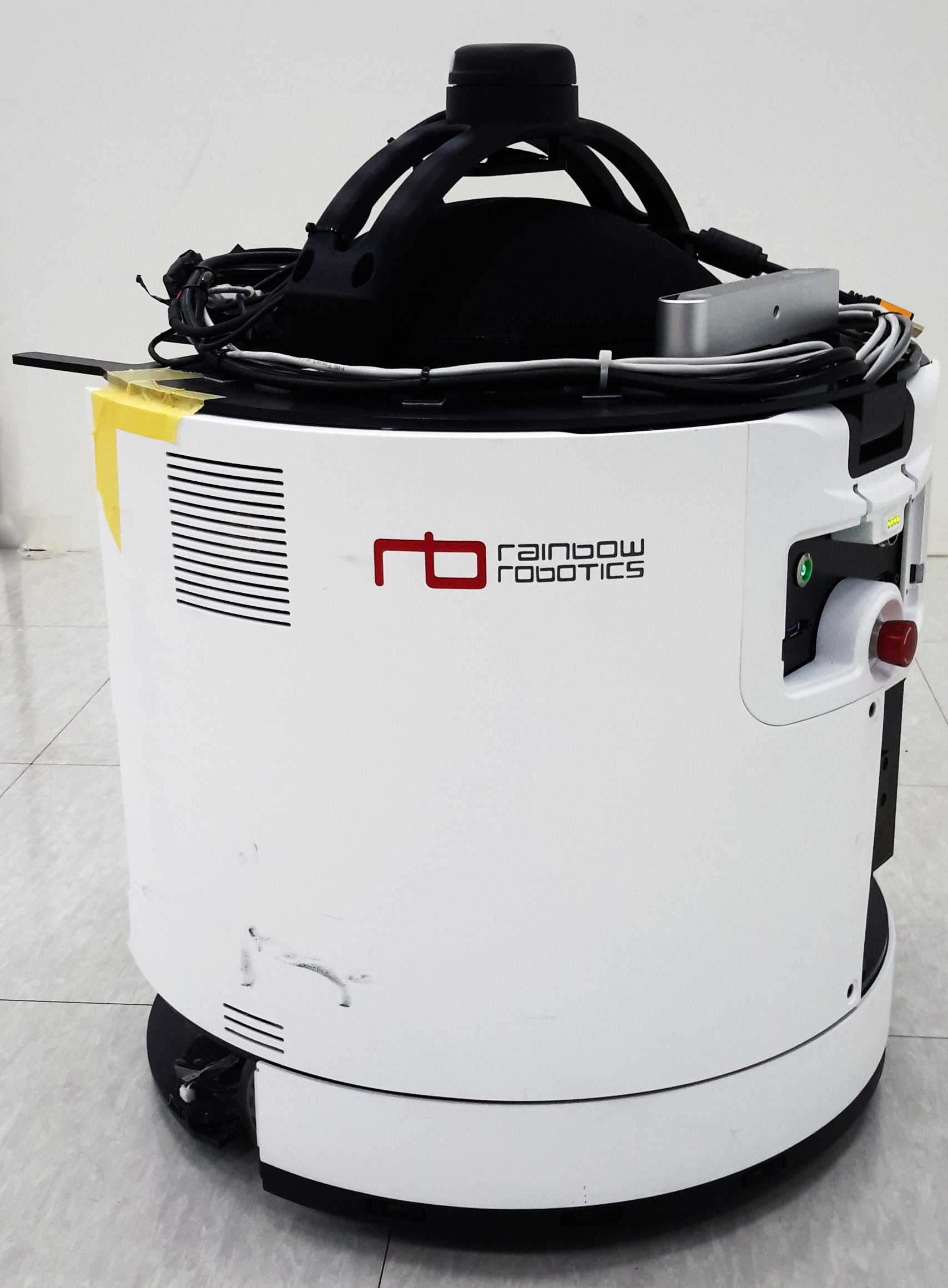}
    }
    \subfigure[Turtlebot with top-mounted 260-degree view \lidar{} \label{fig:turtlebot_intro}]{
    \includegraphics[width=0.4\columnwidth]{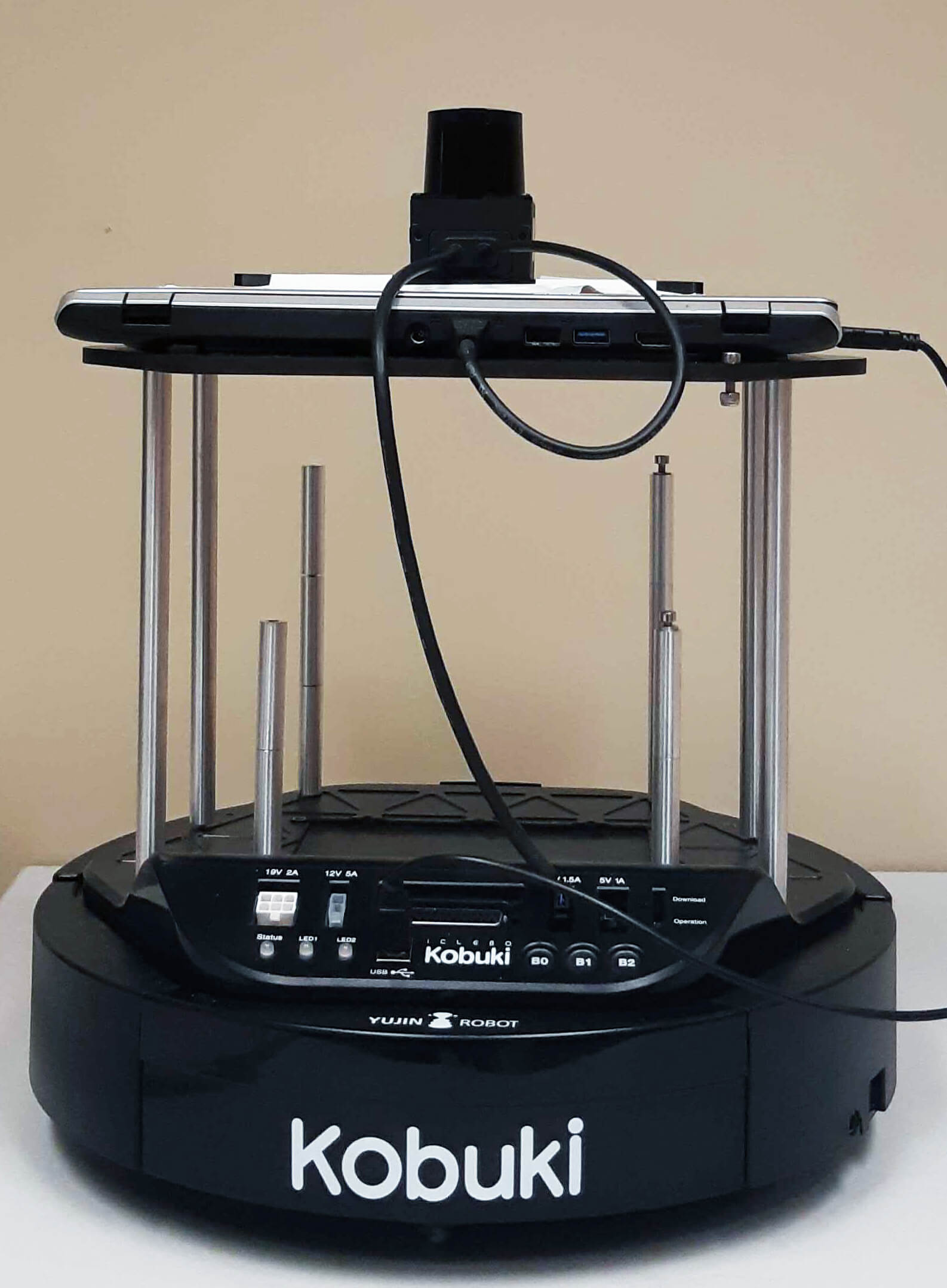}
    }
    \caption{Robots used in our experiments \vspace{-0.3cm}}

\end{figure}

With JAY deployed in the indoor environment shown as `env0' in Fig. \ref{fig:mod_envs}, we tested our likelihood model. 
To compare the localization performance with some of existing methods that run on grids, the experiments included the following methods.
\begin{itemize}
    \item Trained likelihood model (LM): use the trained model for the sensor measurement likelihood at each pose.
    \item Scan Matching (SM): use cosine similarity between the precomputed scan matrix and the input scan for the sensor measurement likelihood.
    \item Reinforcement Learning (RL): use the trained policy model $\pi$ for sampling next action.
    \item Active Markov Localization (AML): the robot virtually goes one step ahead along each of the possible actions and select the one with the largest reward, defined as the decrease in entropy.
    \item Random Action (RA): sample next action randomly with uniform probability.
\end{itemize}



\begin{figure}[!hbt]
    \centering
    \subfigure[Sait map \label{fig:jay_init_poses} ]{
    \includegraphics[width=0.38\columnwidth]{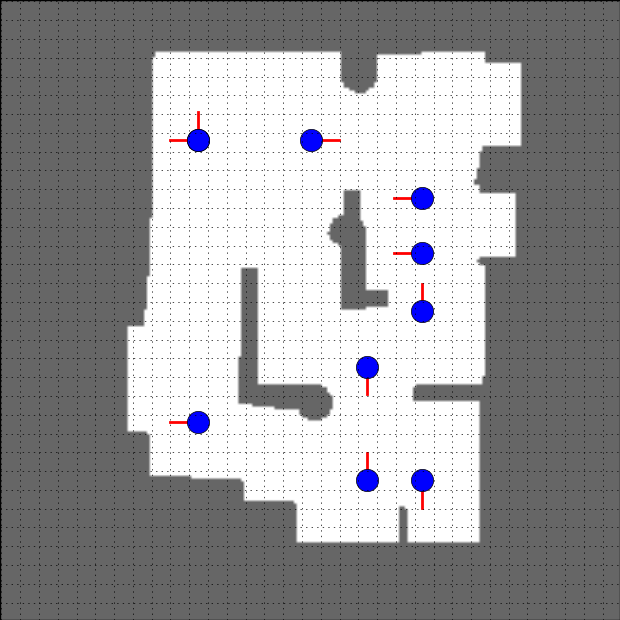}
    }
    \subfigure[Mila map
    \label{fig:mila_map}]{
    \includegraphics[width=0.38\columnwidth]{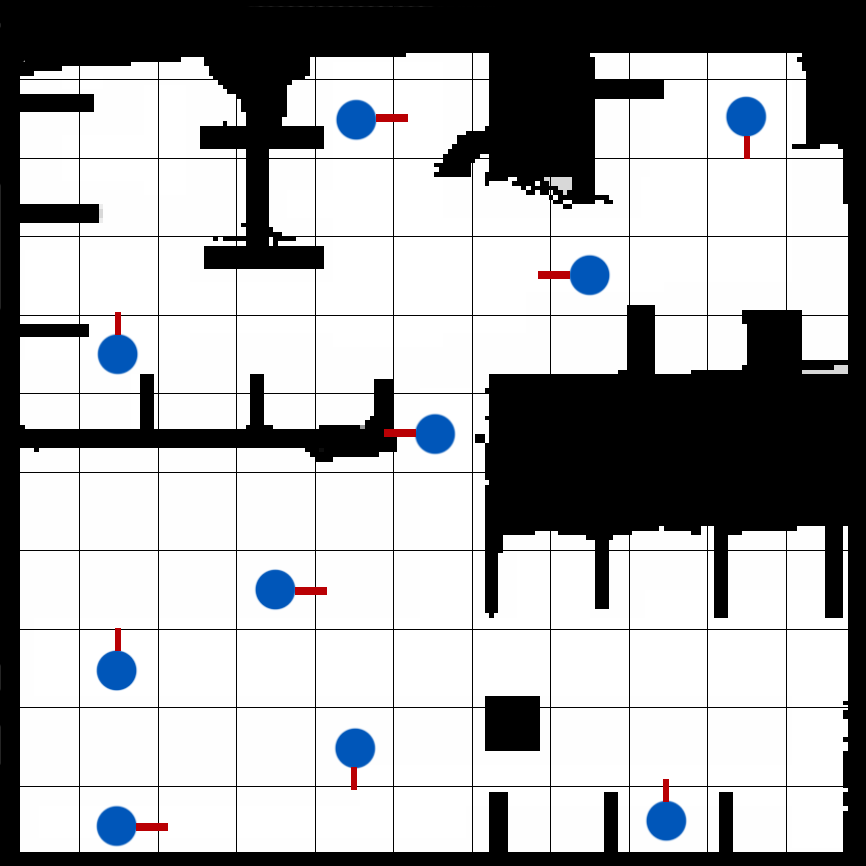}
    }
    \caption{The maps of the real environments in SAIT and Mila were generated by using a SLAM algorithm, \emph{gmapping} \cite{gmapping}. The map size was adjusted to $224\times224$ pixels each of size $0.04m\times0.04m$. We randomly selected 10 poses on the map and used them repeatedly as the starting poses for the 10 episodes of each test. 
    \label{fig:exp_maps}}
\end{figure}

For each test condition, it ran for 10 episodes of length 11. Fig. \ref{fig:jay_init_poses} shows the 10 initial poses that were randomly sampled to be applied to all the tests. 

A custom control law computed velocity command at each step to move JAY to the next target pose by using the odometry feedback, where the target pose was computed from the believed pose $bel{x_t}$ and the action sampled from the policy.

\subsubsection{Results from JAY Experiment}
We evaluated the test results using the following metrics:

\begin{enumerate}
\item[i)] Earth Mover's or Wasserstein's distance $W$: It quantifies the error between our belief map and true belief map (which has probability 1.0 at true pose and 0 everywhere else) 
\begin{equation}
W = \sum_{x} p(x) D(x,x^*),
\end{equation}
where we define $D$ to be the Manhattan distance between two poses.
\item[ii)] estimated belief at true pose (This is also the reward that was used for training) $p(x^*)$
\item[iii)] hit rate, which is defined as the number of times the DAL pose is exactly same as true pose.
\end{enumerate}

\begin{figure}
    \centering
    \subfigure[Wasserstein distance
    \label{subfig:jay_result_11x11}]
    {
    \includegraphics[width=.97\columnwidth]{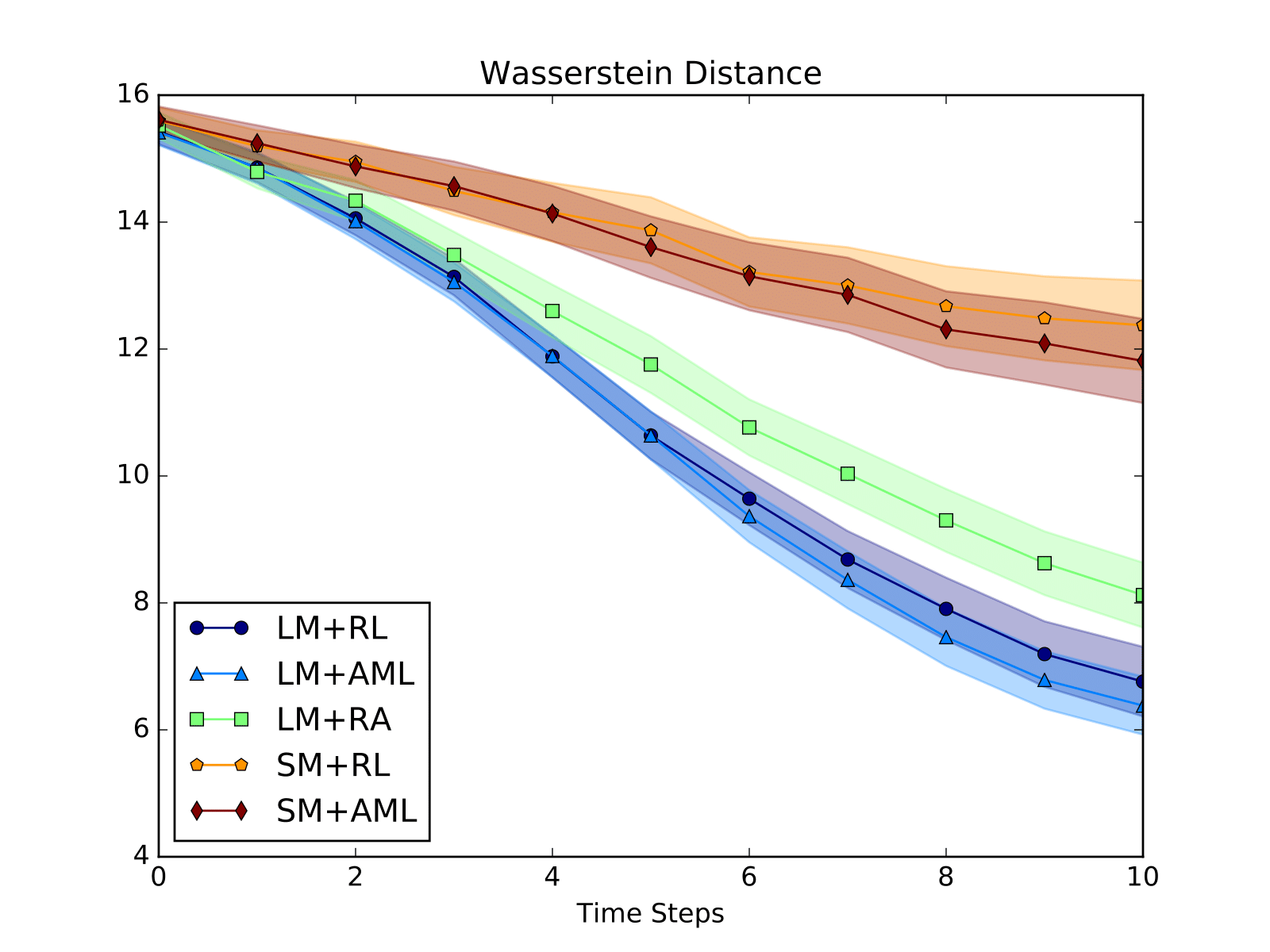} 
    }
    \subfigure[Belief at ground truth
    \label{subfig:jay_result_33x33}]
    {
    \includegraphics[width=.97\columnwidth]{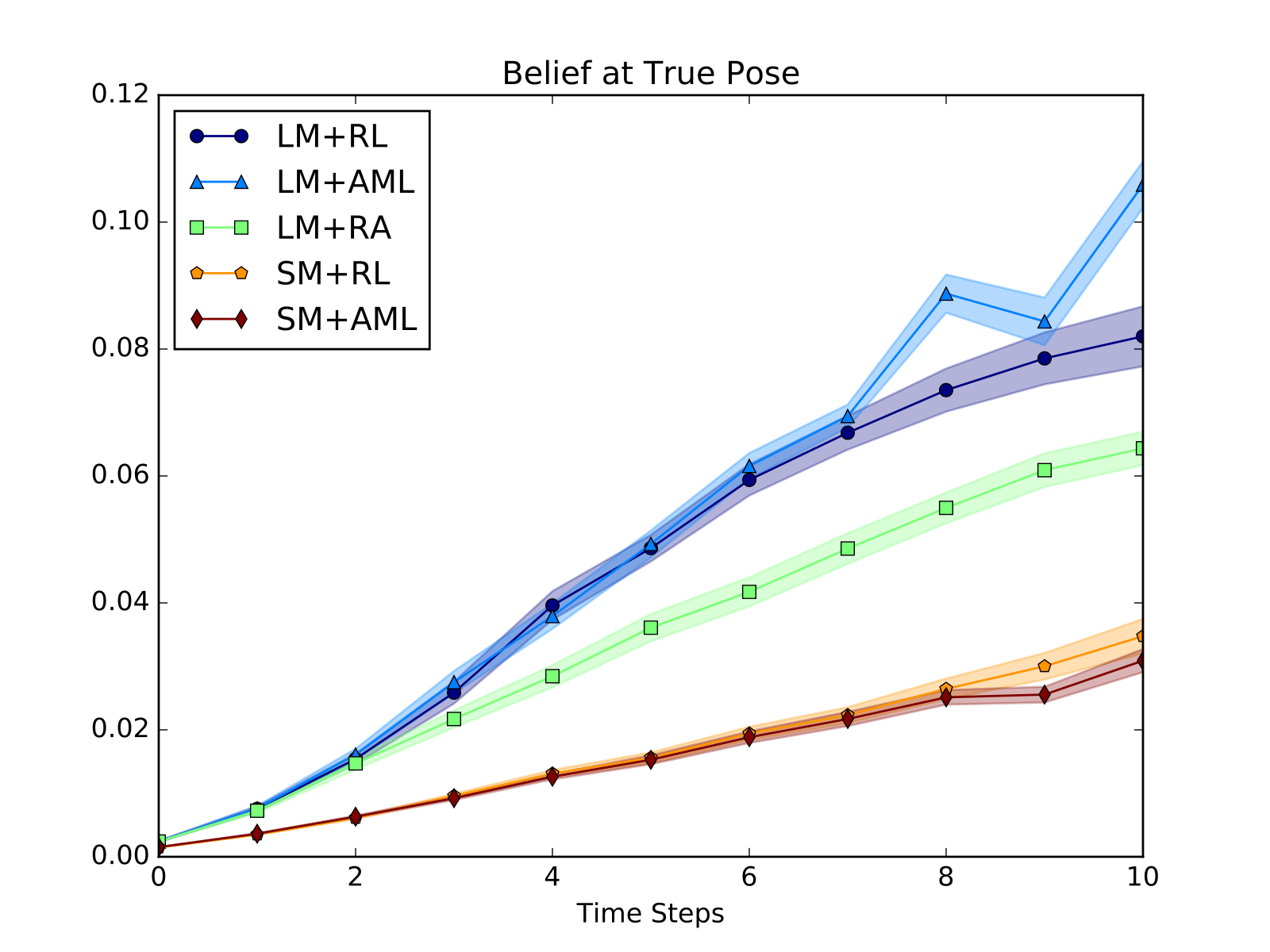} 
    }
    \subfigure[hit rate
    \label{subfig:jay_result_33x33}]
    {
    \includegraphics[width=.97\columnwidth]{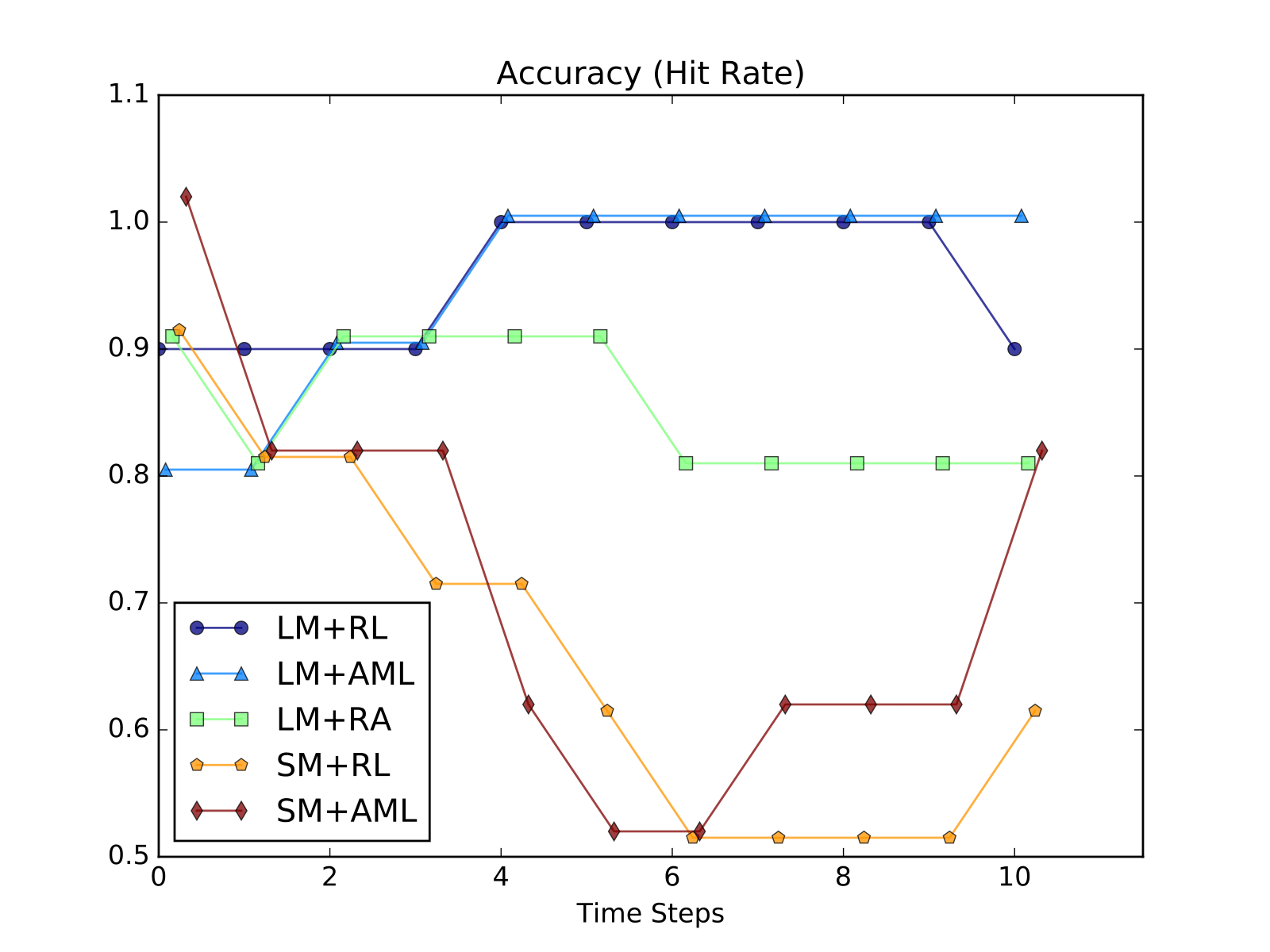} 
    }
    \caption{Localization performance was evaluated in 8 headings and $33 \times 33$ grids with JAY in the real environment. We compared our proposed likelihood model for 33$\times$33 (LM), scan-matching (SM), active Markov localization (AML), random action (RA), and the trained policy $\pi$ with reinforcement learning (RL).}
    \label{fig:jay_exp_results}
\end{figure}

Each test consists of 10 episodes; the mean and standard deviation over the 10 episodes were then computed for each of 11 steps as plotted in \ref{fig:jay_exp_results} to illustrate how the metrics changed during an episode in average. 

From the results plotted in Fig. \ref{fig:jay_exp_results}, we can clearly see that our trained likelihood model performed much better than scan matching in all the metrics. This could be attributed to the susceptibility of SM to noises such as inaccurate map and the robot being off the center of a grid cell. Since we accounted for all these noises during training, our LM proved to be robust. Our RL model performed as good as AML. In terms of time complexity, our RL model is however multiple times faster than AML as we show in the time complexity analysis (cf. Table \ref{table:time-complexity})

\begin{table}[!h]
\centering
\begin{tabular}{|l|l|l|l|l|}
\hline
 & SM & LM & RL & AML(+LM) \\ \hline
4x11x11 & 0.202 & 0.0531 & 0.00122 & 1.23 \\ \hline
4x33x33 & 0.316 & 0.0518 & 0.00172 & 1.29 \\ \hline
8x33x33 & 0.492 & 0.0866 & 0.00203 & 1.38 \\ \hline
24x33x33 & 0.770 & 0.123 & 0.00194 & 1.47 \\ \hline
\end{tabular}
\caption{Time complexity analysis (in seconds) \label{table:time-complexity}
}
\end{table}

\begin{figure}
    \centering
    \subfigure[The original and the 5 modified environments \label{fig:mod_envs}]
    {
        \includegraphics[width=0.95\columnwidth]{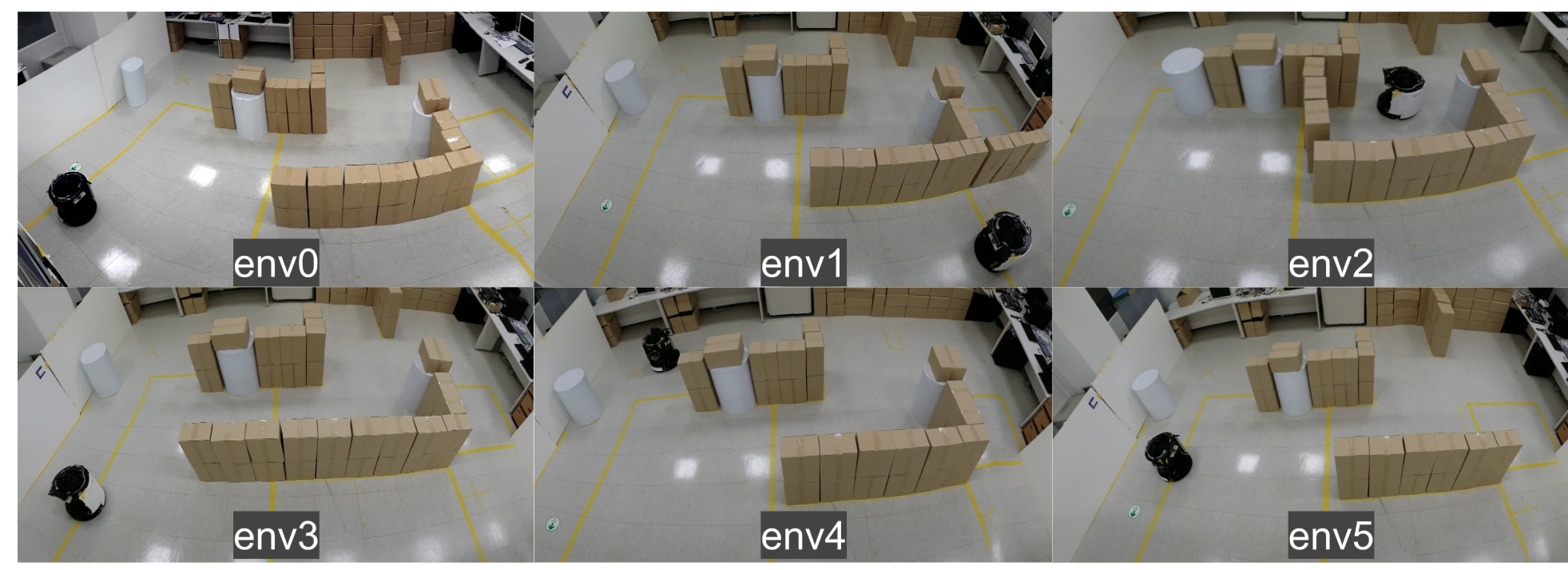}
    }
    \\
    \subfigure[Wasserstein distance between belief and true belief\label{fig:env_changes}]
    {   
        \includegraphics[width=.95\columnwidth]{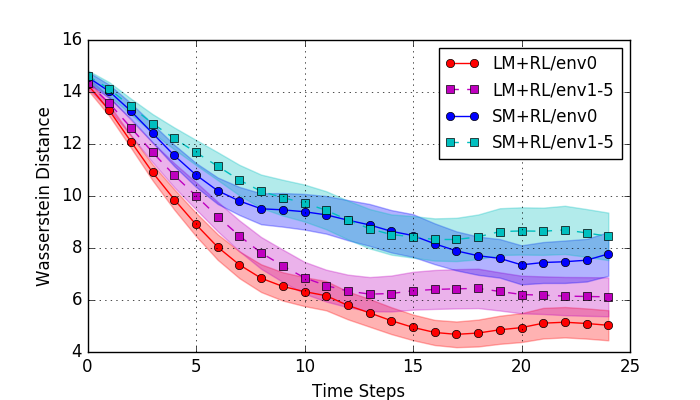} 
    }
    \caption{LM is tested in modified environments env1 to env5 and compared with the error from the original environment env0. Error is measured in terms of Wasserstein distance between belief and true belief. }
    \label{fig:env_changes_sup}
\end{figure}

To further justify our robustness claims of our trained LM, we modified the environment by adding or moving some obstacles and ran the same experiment again.  We tested it for LM+RL versus SM+RL on 5 differently modified environments from `env1' to `env5' shown in Fig. \ref{fig:mod_envs}. We also tested again at the original environment `env0'. Each test was done with the same 10 initial poses as above while the length of an episode was increased to 25. Number of headings was fixed at 4.

\subsubsection{Experiments on Turtlebot}
We used Hukoyo Laser UST-20LX
which outputs 1040 measurement ranges with detection angle of 260° and angular resolution of 0.25°. The netbook runs on Intel Core i3-4010U processor with 4GB RAM. Our test environment is of size $9 \times 9 $ meters$^2$ and is combination of 2 rooms as shown in \ref{fig:mila_map}. Similar results as that of JAY were observed (omitted for brevity). 

\subsection{gym-dal}
Most of the robotics research involves the extensive use of ROS and Gazebo environments which are convenient
but they are not ideal platform for implementing learning algorithms especially reinforcement learning. To alleviate these problems, we are open sourcing our gym environment. We hope that this will act as a testbed for active localization research. Once trained on our simulator, the policies and the likelihood model learned are directly transferable to real robot. 
We have also integrated the environment with various state of the art RL algorithms \cite{iko} like PPO \cite{PPO}, A2C\cite{Mnih2016}, ACKTR \cite{ACKTR}.

\subsection{Hierarchical Likelihood models}

\begin{figure}[!hbt]
    \centering
    \subfigure[loss at level-0
    \label{fig:hle_loss}]{
    \includegraphics[width=0.4\columnwidth]{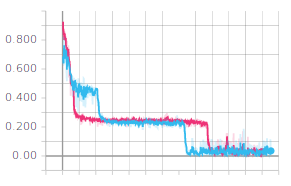}
    }
    \subfigure[loss at level-1]{
    \includegraphics[width=0.4\columnwidth]{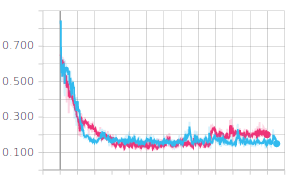}
    }
    \caption{blue: weighted HLE. red: unweighted HLE}

\end{figure}

We present the experimental results of our 2 novel hierarchical likelihood models. From Fig.~\ref{fig:hle_loss}, we can observe that both weighted and un-weighted variants of Hierarchical likelihood model performed equally well and obtained convergence. However, higher levels of hierarchy are very sensitive to hyper-parameters because they are dependent on the performance of previous layers. The robustification studies of the hierarchical model, a thorough empirical evaluation and making them insensible to hyper parameters are possible avenues for future research.

\section{Related Work}
\label{sec:related}


Localization is an extremely well-studied robotics problem. We present a treatment of relevant \emph{active} and \emph{passive} approaches to robot localization, drawing parallels and contrast to our method.

\subsection{\emph{Passive} localization}
\label{sec:pass_loc}
\emph{Passive} localization methods can broadly be classified into two representative categories: \emph{Bayesian} filtering methods and learning-based methods. Although related, we do not cover visual place recognition approaches as their primary intention is to provide \emph{coarse} location estimates. We refer the interested reader to \cite{place_rec_survey} for an excellent survey.

\subsubsection*{\textbf{Bayesian filtering-based methods}}
\label{sec:Bayes_filt}
Off-the-shelf localization modules such as Extended Kalman Filter (EKF)-based localization \cite{probabilistic_robotics}, Markov localization \cite{markov_localization} and Adaptive Monte-Carlo localization (AMCL) \cite{amcl} are based on a Bayesian filtering viewpoint of localization. All these methods begin with a prior (usually uniform) \emph{belief} of a robot's pose and recursively update the belief as new (uncertain) measurements arrive. The representation of the belief is what fundamentally differentiates each approach. EKF-based methods assume a Gaussian distribution to characterize the belief, while Markov localization \cite{markov_localization} and AMCL \cite{amcl} use non-parameteric distributions to characterize the belief (histograms and particles respectively). Consequently, EKF localization lacks the capability to characterize multiple belief hypotheses (owing to the unimodal Gaussian prior), while Markov localization and AMCL suffer computational overhead as the population of the non-parameteric distribution increases.

\subsubsection*{\textbf{Learning-based methods}}
\label{sec:learning_meth}
Recently there has been a resurgence of learning-based methods for localization. Approaches such as PoseNet \cite{posenet} and VLocNet \cite{Valada18icra} perform \emph{visual} localization by training a convolutional neural network to regress to \emph{scene coordinates}, given an image \cite{posenet} or a sequence of images\cite{Valada18icra}. Recently, differentiable particle filters (DPFs) have been proposed for global localization \cite{DPF1,DPF2}. However, such approaches need precisely annotated data to train a \emph{PoseNet}, \emph{VLocnet}, or a \emph{DPF} for each new deployed environment. In contrast, DAL transfers essentially \emph{zero-shot} across simulated environments, and from a simulator to the real-world.

\subsection{\emph{Active} localization}
\label{sec:active_loc}

In terms of \emph{active} localization methods, there are again two broad categories: information-theoretic approaches and learning-based approaches.

\subsection*{\textbf{Information theoretic methods}}
\label{sec:info_theo}

Burgard et al. introduced \emph{active} localization in their seminal work \cite{active_seminal}. They demonstrated that, rather than passively driving a robot around, picking actions that reduce the expected localization uncertainty results in a more efficient and robust solution. Using an entropy measure characterized as a mixture-of-Gaussians, they demonstrate that the framework of Markov localization \cite{markov_localization} can be extended to action selection. This was more a \emph{proof-of-concept}, as the applicability of this method is confined to low-dimensional state-spaces, where entropy computation can be carried out efficiently. Since then, several other approaches \cite{Feder_IJRR_1999,Roy_ICRA_1999,Mariottini_ICRA_2011,Valencia_IROS_2012,Forster_RSS_2014,Kim_IJRR_2015} use a similar, information gain maximization cost for the task of active localization or SLAM.

\subsection*{\textbf{Learning-based methods}}
\label{sec:lear_meth_act}

Active Neural Localization (ANL) \cite{ANL} is the first known work to employ a learned model for active localization from images. Their approach comprises two modules: a perceptual model and a policy model. 
The perceptual model is not completely learned. The transition functions here are assumed to be deterministic, and this prohibits the policy model to be transferred onto a real robot. On the other-hand, DAL comprises a likelihood and a policy model, both of which are learned from data, which outperform their traditional counterparts. The hierarchical likelihood estimation also allows for scalability to much bigger environments compared to ANL.

\section{Conclusion} 
\label{sec:conclusion}

We have presented a learning-based approach to active localization from a known map. We propose multi-scale learned perceptual models that are connected with an RL planner and inner loop controller in an end-to-end fashion. We have demonstrated the effectiveness of the approach on real robot settings in two completely different setups.

In future work, we would like to explore the removal of the reliance of the system on a high fidelity map such as is generated by \emph{gmapping}. Much more appealing would be, for example, if the robot could localize on a hand drawn sketch or some much more easily obtained representation.  


\bibliographystyle{plain}
\bibliography{dalbib}

\end{document}